\documentclass[10pt,letterpaper]{article}
\usepackage[margin=.75in,columnsep=.25in]{geometry}
\usepackage{newtxtext,newtxmath}
\usepackage{graphicx,booktabs,microtype,xspace,enumitem,pifont}
\usepackage[numbers,sort&compress]{natbib}
\usepackage{titlesec}
\usepackage{caption,multicol}
\usepackage[colorlinks=true,allcolors=blue]{hyperref}
\titleformat{\section}{\large\bfseries\centering}{\thesection}{.6em}{}
\titleformat{\subsection}{\normalsize\bfseries}{\thesubsection}{.6em}{}
\graphicspath{{figures/}}

\newcommand{\rsgpo}{\textsc{RobustSGPO}\xspace}

\hypersetup{pdftitle={RobustSGPO: Search-Space Control for Agent Harness Evolution},pdfauthor={Zibo Zhao; Jijun Shi; Mo Zhou; Zhongyuan Wang; Shifu Bie; Yunfei Zhang; Xuanting Zhou; Xiangyu Wu; Bin Liu; Ruiming Tang; Wenwu Ou; Kun Gai}}
\begin{document}
\noindent\includegraphics[width=1.05in]{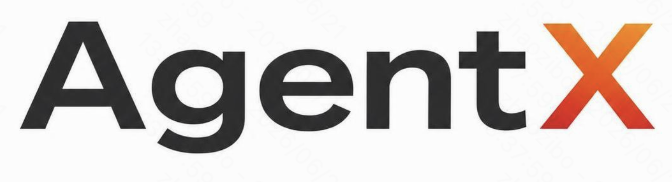}\par
\vspace{7pt}
\begin{center}
{\fontsize{14}{16}\selectfont\bfseries
RobustSGPO: Search-Space Control for Agent Harness Evolution\par}
\vspace{12pt}
{\fontsize{12}{14}\selectfont\bfseries
Zibo Zhao\textsuperscript{1,2,\textdagger}, Jijun Shi\textsuperscript{2}, Mo Zhou\textsuperscript{2}, Zhongyuan Wang\textsuperscript{2}, Shifu Bie\textsuperscript{2}, Yunfei Zhang\textsuperscript{2},\\
Xuanting Zhou\textsuperscript{2}, Xiangyu Wu\textsuperscript{2}, Bin Liu\textsuperscript{2}, Ruiming Tang\textsuperscript{2}, Wenwu Ou\textsuperscript{2}, Kun Gai\textsuperscript{2}\par}
\vspace{6pt}
{\fontsize{10}{11}\selectfont
\textsuperscript{1}Wuhan University, Wuhan, China\qquad 
\textsuperscript{2}Kuaishou Technology, Beijing, China\par}
\vspace{2pt}
{\fontsize{9}{10}\selectfont
\mbox{whubear@whu.edu.cn; \{shijijun, zhoumo, wangzhongyuan03, bieshifu03, zhangyunfei05, zhouxuanting05, wuxiangyu06\}@kuaishou.com}\par}
\end{center}
\vspace{12pt}
\begin{multicols}{2}\raggedcolumns
\begin{abstract}

Semantic-gradient-based prompt optimization (SGPO) improves agent harnesses using execution feedback, but its local update rule leaves the choice of edit scope and operation unresolved. We introduce \rsgpo, which specifies the requested edit, constructs and checks the patch, and continues search from either the incumbent or retained snapshots. We evaluate permission scheduling, cumulative controls, and task-family transfer in the AgentX brainstorming workflow using 120 tasks, 95 runs, and 7,350 candidate attempts. Periodic $1\to2\to3$ scheduling exceeds fixed maximum permission by 0.28 test-score points. RobustSGPO increases completion on 30 held-out tasks from 60.0\% to 80.0\% and improves test quality from 3.77 to 4.14 under a 20-million-token budget. Category retention reduces source-task degradation after a shift, whereas random retention reaches a higher destination endpoint. Search-space control benefits quality through executable edits and alternative starting points, with measurable retention overhead.

\end{abstract}
\section{Introduction}
This paper builds on the published AgentX work~\cite{agentx2026}, which established an industrial recommendation workflow and a semantic-gradient-based prompt optimization (SGPO) loop for harness evolution. SGPO diagnoses trace failures, edits an agent specification, and admits candidates through paired replay. RobustSGPO extends this foundation with explicit search-space control.

Multi-agent behavior is distributed across instructions, input/output contracts, and agent connections. Restricting edits to one agent limits direct repair of cross-agent failures. Opening the whole permitted space does not ensure that the model explores it: generation can still concentrate on familiar instruction rewrites. We study how to select edits, make them executable, and retain alternative starting points under a fixed budget.

Our contributions are an analysis of the gap between permitted and valid realized edits; RobustSGPO's explicit control of edit selection, execution, and retention; and experiments connecting these controls to held-out quality, transfer, and token cost.

\section{Related Work}
\textbf{Harness evolution.} Meta-Harness searches code using prior candidates, scores, and traces~\cite{metaharness2026}; AHE exposes editable components and links changes to predicted outcomes~\cite{ahe2026}. Self-Harness couples failure mining with minimal proposals and regression validation~\cite{selfharness2026}. These works establish harness evolution beyond prompt rewriting. RobustSGPO studies a narrower question: selecting scope, operation, and targets before generation, then enforcing targets and constructing prescribed add/remove edits.

\textbf{Structured search and adaptation.} AgentFlow searches roles, topology, and protocols through a typed graph DSL with pre-execution checks~\cite{agentflow2026}. HARBOR searches bounded harness configurations with cost-aware Bayesian optimization~\cite{harbor2026}. HELIX provides typed, source-traceable composition and verified trajectories for model--harness co-evolution~\cite{helix2026}; HASE jointly evolves solutions, harnesses, and model weights~\cite{hase2026}. Adaptive Auto-Harness routes tasks through a harness tree~\cite{adaptiveharness2026}, while Evo-Harness compiles experience into reusable skills~\cite{evoharness2026}. Our model and evaluator stay fixed; archived snapshots seed further mutations and must pass incumbent-based admission, rather than serving as task-time routes.

\textbf{Prompt evolution and retention.} OPRO, Promptbreeder, and TextGrad optimize textual variables~\cite{opro2023,promptbreeder2023,textgrad2024}; GEPA retains complementary candidates on a Pareto frontier~\cite{gepa2025}. Our MAP-Elites-inspired archive~\cite{mapelites2015} instead groups complete snapshots by edit scope and operation. The experiments isolate these cumulative controls within AgentX; they do not rank RobustSGPO against the above systems on shared benchmarks. Evaluation covers one brainstorming workflow; fixed tables and code reproduce its curves.

\end{multicols}
\clearpage
\begin{multicols}{2}\raggedcolumns
\section{System Context and Baseline SGPO}
\subsection{AgentX and the Evaluation Workflow}
In the AgentX workflow (Figure~\ref{fig:agentx}), the Brainstorm Agent generates experimental ideas; its Developing Agent changes code; its Evaluation Agent analyzes experiments; harness evolution improves these agents~\cite{agentx2026}. We evaluate the brainstorming workflow, in which proposal orchestration coordinates question, idea, and validation agents. Transfer stays within this workflow.

\end{multicols}
\par\addvspace{8pt}\noindent
\begin{minipage}{\textwidth}

  \centering
  \includegraphics[width=.80\textwidth]{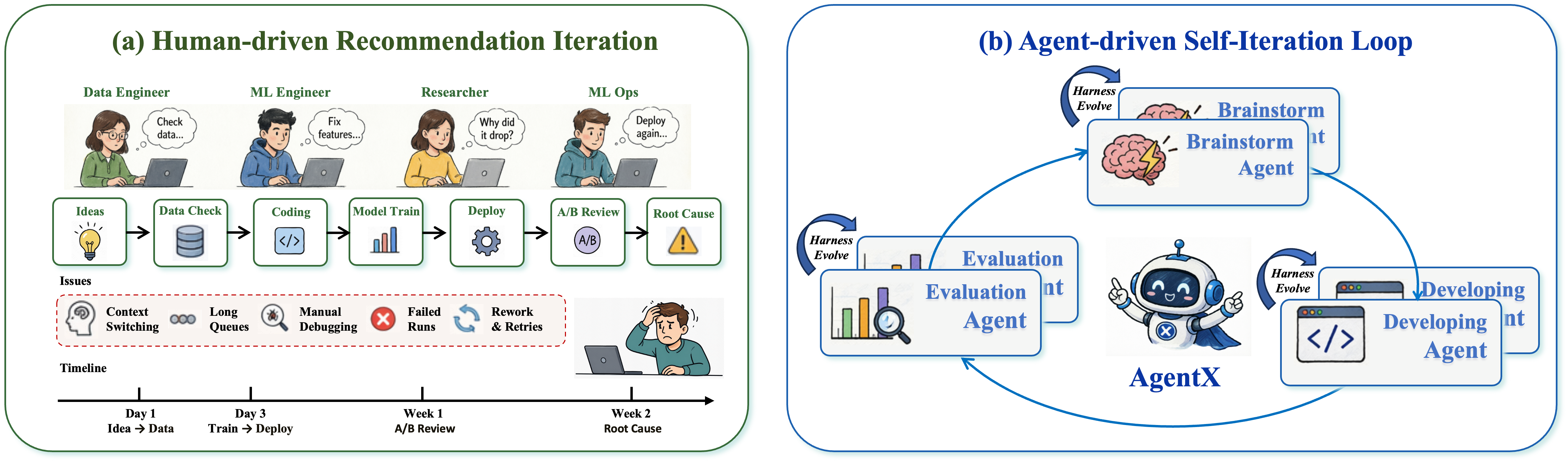}
  \captionof{figure}{AgentX changes industrial recommendation iteration from a manually
  handed-off pipeline into an agent-driven closed loop. Brainstorming,
  development, and evaluation agents consume online feedback and trajectory data,
  while harness evolution improves each agent over time. Prior-work foundation, reproduced from the
  AgentX technical report \cite{agentx2026}.}
  \label{fig:agentx}

\end{minipage}
\par\addvspace{12pt}
\begin{multicols}{2}\raggedcolumns

\subsection{Trace-to-Update Loop}
The report's SGPO-I updates one target agent $i$, keeping the model, tools, orchestration, and other agents fixed~\cite{agentx2026}. It samples traces $\mathcal{T}$, derives rubrics $\mathcal{R}$ and standalone replay tasks, and computes a natural-language loss and semantic gradient:
\begin{equation}
\ell_{t,i},g_{t,i}=E_{\mathrm{agent}}(h_{t,i};\mathcal{T},\mathcal{R}).\label{eq:sgpo-gradient}
\end{equation}
The gradient identifies missing requirements, sequencing errors, or broken contracts. The refinement agent generates a candidate:
\begin{equation}
h'_{t,i}=R_{\mathrm{agent}}(h_{t,i},g_{t,i}).\label{eq:sgpo-refine}
\end{equation}
The experiment agent replays identical tasks on both versions. With $\Delta J_i=\operatorname{ReplayScore}(h'_{t,i})-\operatorname{ReplayScore}(h_{t,i})$,
\begin{equation}
h_{t+1,i}:=h'_{t,i}\quad\text{if }\Delta J_i>\epsilon\land\operatorname{Safe}(\Delta h_i).\label{eq:accept}
\end{equation}
Otherwise the update is rolled back and its patch, score, and diagnosis enter refinement experience. Rubrics, verifiers, traces, and model paths are protected. Figure~\ref{fig:sgpo} reproduces this loop.

\end{multicols}
\par\addvspace{8pt}\noindent
\begin{minipage}{\textwidth}

  \centering
  \includegraphics[width=.74\textwidth]{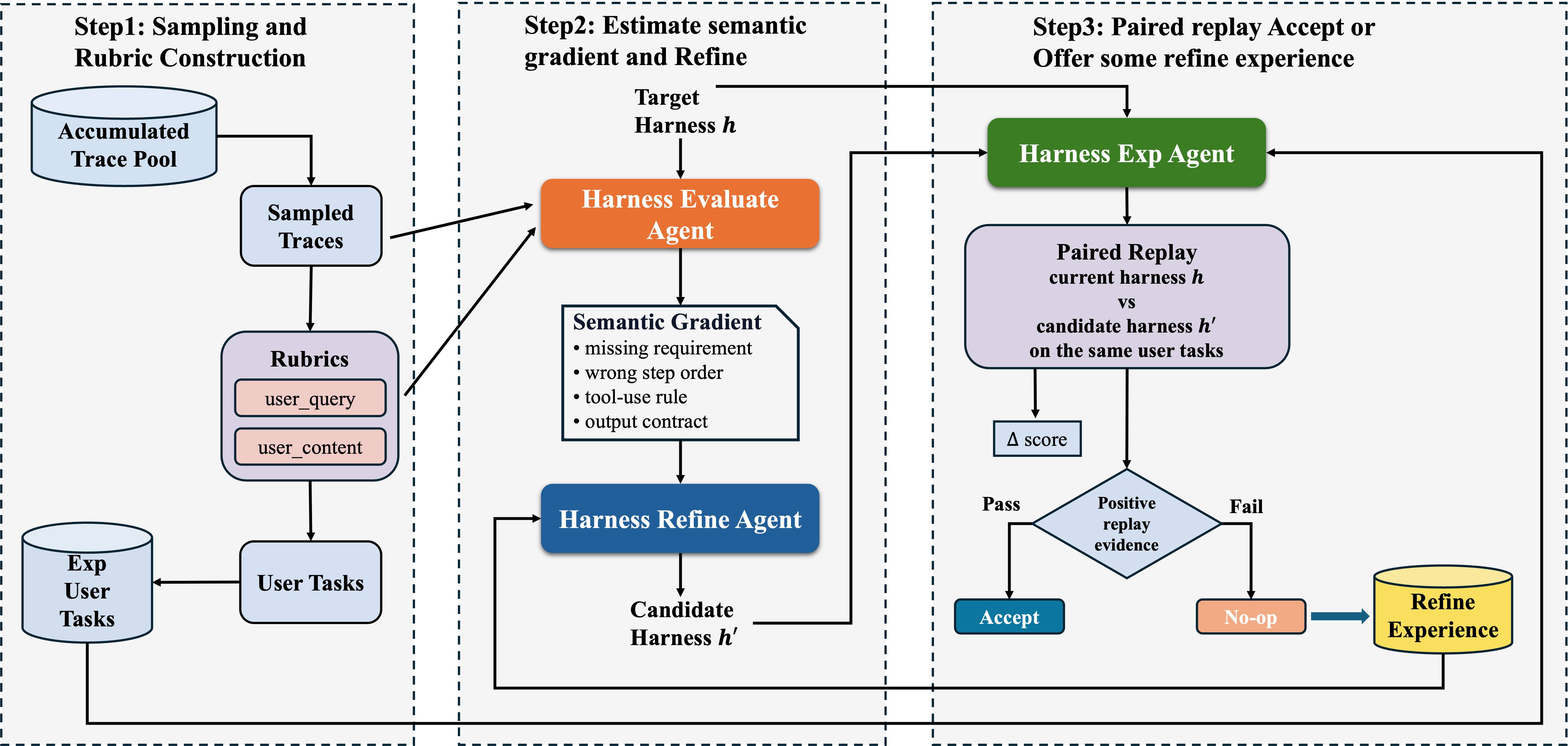}
  \captionof{figure}{Prior-work SGPO-I loop: trace sampling and task construction, semantic-gradient refinement, and paired replay with accept/rollback. Reproduced from the AgentX technical report~\cite{agentx2026}.}
\label{fig:sgpo}

\end{minipage}
\par\addvspace{12pt}
\begin{multicols}{2}\raggedcolumns

\section{The Search-Space Problem}
The nominal space $\mathcal{S}_{\rm nom}$ contains permitted edits; the realized space $\mathcal{S}_{\rm real}$ contains actually generated edits. Validity checks select the effective space:
\begin{equation}
\mathcal{S}_{\rm eff}=\{\Delta H\in\mathcal{S}_{\rm real}:C_{\rm path}\land C_{\rm semantic}\land C_{\rm budget}\land C_{\rm safe}\}.
\end{equation}
Effective means valid, not admitted by replay. Broader permissions need not yield more valid edits or higher quality. We use three levels: $\alpha_1$ edits one designated agent; $\alpha_2$ edits any existing brainstorming agents but preserves their set; $\alpha_3$ additionally permits agent addition/removal and routing changes.

E1 compares permission schedules; E2 measures valid-category coverage and structural validity, not all three space sizes. E3 tests retained starting points after a task shift.

\section{Robust SGPO}
RobustSGPO extends AgentX's SGPO while retaining its diagnosis, proposal, replay, and accept/rollback loop. It makes three decisions explicit: what edit to attempt, how to construct and check it, and which version to continue from. A running AgentX example explains these decisions below; Figure~\ref{fig:robust} locates them within the original loop.

\end{multicols}
\par\addvspace{8pt}\noindent
\begin{minipage}{\textwidth}

\centering
\includegraphics[width=.74\textwidth]{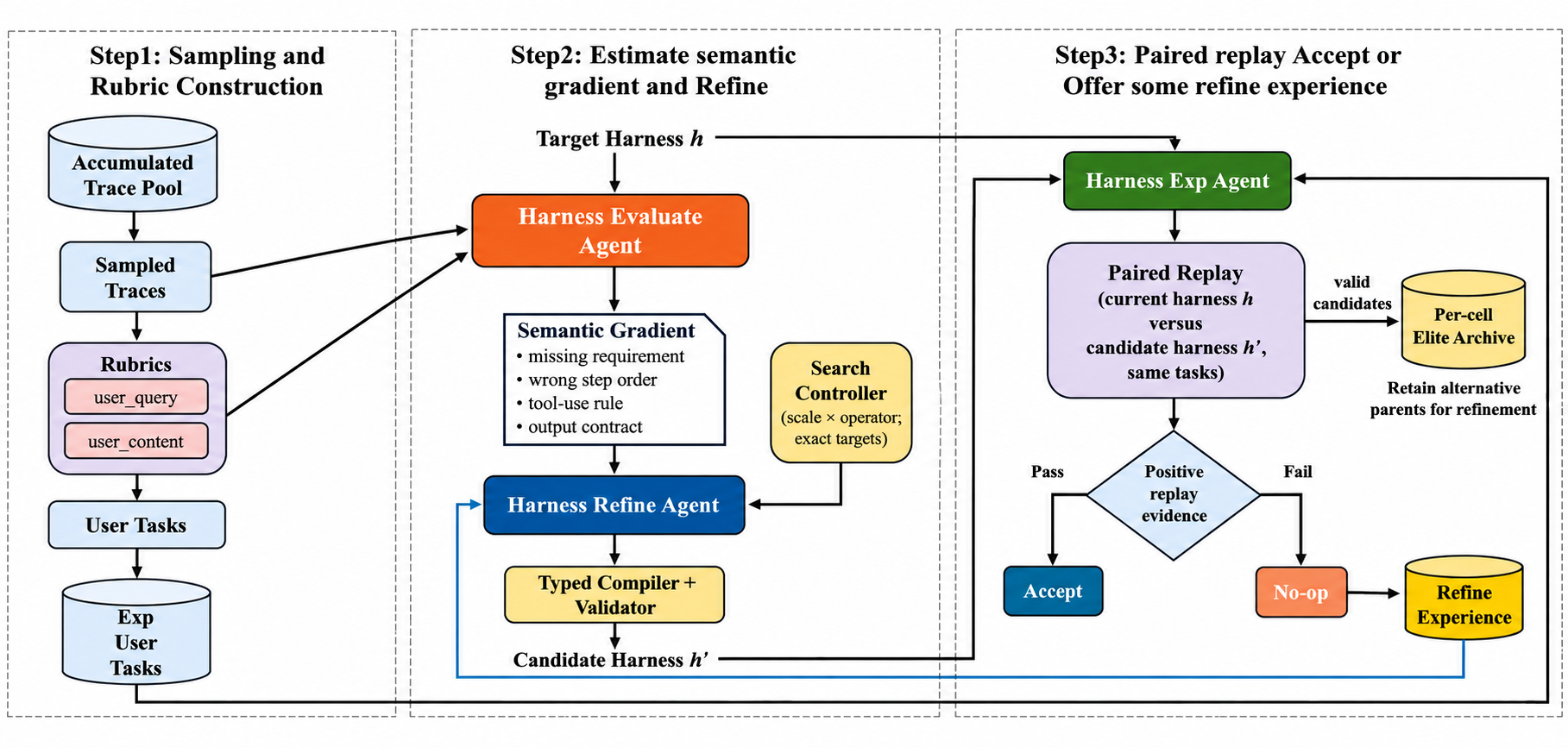}
\captionof{figure}{RobustSGPO extends Figure~\ref{fig:sgpo} with edit selection, patch construction/checking, and retained search starting points (Sections~\ref{sec:factorized}--\ref{sec:retained}). All candidates still undergo incumbent-based replay admission.}
\label{fig:robust}

\end{minipage}
\par\addvspace{12pt}
\begin{multicols}{2}\raggedcolumns

\subsection{Choose What to Change This Round}\label{sec:factorized}
Consider an illustrative AgentX failure: \texttt{single-idea} supplies only a vague evidence claim, but \texttt{single-validate} needs its source and finding. A freely generated repair may merely tell idea to provide more evidence. RobustSGPO instead specifies: change both agents' output/input requirements to use matching fields. The model then writes the concrete instructions.

This request selects scope $s$ (a pair), operation $o$ (field alignment), and targets $T$ (idea and validate). The controller chooses feasible categories $c=(s,o)$ and eligible targets before content generation. Given feedback $g$,

\end{multicols}
\par\addvspace{8pt}\noindent
\begin{minipage}{\textwidth}

\begin{equation}
q(\Delta h\mid h,g)=\sum_{s,o,T}p(s,o\mid h)\,p(T\mid s,o,h)\,q_{\rm LLM}(\Delta h\mid h,g,s,o,T).
\end{equation}

\end{minipage}
\par\addvspace{12pt}
\begin{multicols}{2}\raggedcolumns

Only feasible combinations are included; scope and operation need not be independent. Permissions bound allowed changes; factorization directs realized edits within that boundary.

\subsection{Construct and Check the Requested Patch}
The exact-target check compares requested and changed object sets. If the request names both idea and validate, changing only idea or also changing propose is rejected. This filters realized edits for object compliance; it does not establish semantic correctness or task quality.

For additions/removals, predefined code constructs the patch. Given new-agent content, it writes the prescribed file and normalizes an existing name field; removal deletes the designated file. A free-form model could write code to do the same, but a checker would only accept or reject its output. The predefined operation performs the repeated editing work itself, helping requested structural edits pass validity checks. Ordinary instruction edits remain model-generated. This is the Typed Compiler in Figure~\ref{fig:robust}.

The current program does not automatically update caller instructions: creating an evidence agent does not itself insert it between idea generation and review. Loading, references, invocation on trigger tasks, and task quality are checked separately. Candidate admission retains Equation~\ref{eq:accept}; failures also consume budget.

\subsection{Continue from Retained Versions}\label{sec:retained}
Suppose an incumbent scores 4.2. A local rewrite A scores 4.3 and is accepted; a pair edit B scores 4.1 after fixing the handoff but making review too strict. These numbers illustrate the selection rule. RobustSGPO can retain B as its category's best valid snapshot and later refine its review rule. Any descendant must beat the current incumbent, not merely B.

The archive $A[c]$ stores the highest-validation complete harness snapshot for each retained scope--operation category. Membership follows the requested and verified edit. A seeded reservoir selects at most ten category slots; within retained categories, better candidates replace earlier entries. Search alternates incumbent and archive parents, selecting occupied categories uniformly and falling back to the incumbent when needed. Task shifts trigger rescoring before reuse, charged to search cost. This borrows category retention from quality-diversity search~\cite{mapelites2015}, but categorizes edit actions rather than behavioral niches. Retention changes future search parents, not edit permissions.

\section{Experimental Results}\label{sec:evaluation}
\subsection{Tasks, Protocol, and Runs}
We use 120 tasks in two equally sized families: local instruction failures and cross-agent handoff/structure failures. Grouping by source conversation and underlying requirement precedes a 30/15/15 optimization/validation/test split per family, yielding 60/30/30 tasks. Candidate generation receives optimization traces and static edit errors; validation scores guide selection and retention, while test feedback is excluded from search.

Table~\ref{tab:budget} summarizes runs; all conditions share the initial harness, model, rubric, and proposal budget. We use five paired seeds and three proposals per round, capped at 120 changed lines and 6,000 added characters per proposal. Invalid proposals and retries consume budget. Three paired replays compare candidates and incumbents on identical tasks and execution seeds. Admission requires a mean validation gain above $\epsilon=0.05$ and passing safety checks; at most one best valid candidate is accepted per round, with ties retaining the incumbent. Token costs include generation, failures, replay, judging, and archive operations. Fixed test evaluation is separate from search cost.

\end{multicols}
\par\addvspace{8pt}\noindent\begin{minipage}{\textwidth}
\centering\small
\captionof{table}{Runs and candidate attempts. All conditions use five seeds; E3 includes two transfer directions and random retention.}
\label{tab:budget}
\begin{tabular}{lrrrr}
\toprule
Study & Conditions & Runs & Rounds & Attempts\\
\midrule
E1: Scheduling & 6 & 30 & 30 & 2,700\\
E2: Ablations & 5 & 25 & 30 & 2,250\\
E3: Transfer & $4\times2$ & 40 & 20 & 2,400\\
\midrule
Total & & 95 & & 7,350\\
\bottomrule
\end{tabular}
\end{minipage}
\par\addvspace{12pt}
\begin{multicols}{2}\raggedcolumns

\subsection{E1: Periodic Scheduling Improves Final Quality}
We compare fixed permissions, two periodic schedules, and a shuffle with ten rounds per permission. All use greedy admission without an archive. In Figure~\ref{fig:schedule-plan}, final scores for periodic $1\to2\to3$, periodic $3\to2\to1$, and random scheduling are 4.34, 4.24, and 4.17; fixed $\alpha_1,\alpha_2,\alpha_3$ attain 3.85, 4.07, and 4.06. Fixed $\alpha_3$ leads early, but periodic $1\to2\to3$ finishes 0.28 points higher.

\end{multicols}
\par\addvspace{8pt}\noindent
\begin{minipage}{\textwidth}

\centering
\includegraphics[width=.96\textwidth]{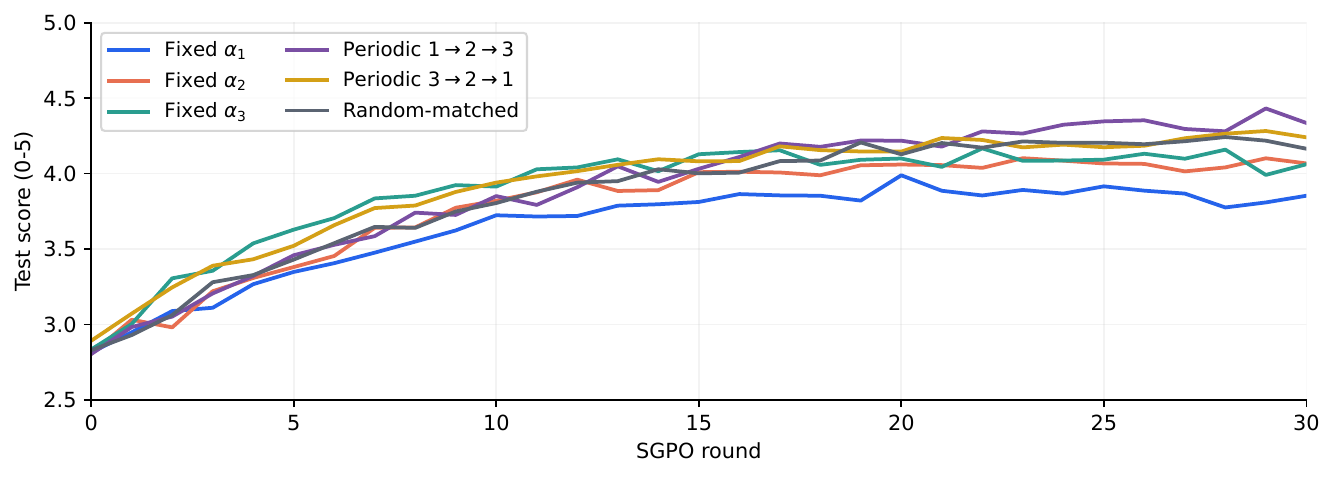}
\captionof{figure}{E1 test-score trajectories on a 0--5 scale. Periodic $1\to2\to3$ has the highest final score; early and final rankings differ.}
\label{fig:schedule-plan}

\end{minipage}
\par\addvspace{12pt}
\begin{multicols}{2}\raggedcolumns

\subsection{E2: Cumulative Search-Space Controls}
Table~\ref{tab:controls} adds a requested edit, exact-target rejection, predefined agent additions/removals, and category retention. All configurations use $\alpha_3$ permission and shared safety/replay checks. Controllers uniformly sample feasible categories and eligible targets. Intermediate controls use matched requests. RobustSGPO alternates incumbent and retained parents; every candidate must beat the current incumbent for admission.

\end{multicols}
\par\addvspace{8pt}\noindent
\begin{minipage}{\textwidth}

\centering\small
\captionof{table}{Cumulative configurations: each row retains all preceding controls. Dashes denote absent additional controls; safety and paired replay apply to every row.}
\label{tab:controls}
\renewcommand{\arraystretch}{1.12}
\begin{tabular}{lcccc}
\toprule
Configuration & Edit plan & Target enforcement & Fixed structural editing & Category archive\\
\midrule
\textbf{SGPO} & --- & --- & --- & ---\\
\textbf{Planned Search} & \ding{51} & --- & --- & ---\\
\textbf{Constrained Search} & \ding{51} & \ding{51} & --- & ---\\
\textbf{Structured Search} & \ding{51} & \ding{51} & \ding{51} & ---\\
\textbf{RobustSGPO} & \ding{51} & \ding{51} & \ding{51} & \ding{51}\\
\bottomrule
\end{tabular}

\end{minipage}
\par\addvspace{12pt}
\begin{multicols}{2}\raggedcolumns

\textbf{Quality and coverage.} Figure~\ref{fig:mechanism-plan}(a) reports round-30 scores of 3.82, 3.94, 4.04, 4.20, and 4.30 in table order: RobustSGPO improves on SGPO by 0.48 points. Final effective-category coverage in Figure~\ref{fig:mechanism-plan}(b) is 48\%, 46\%, 48\%, 54\%, and 68\%, respectively. Providing a request alone does not increase final effective coverage; prescribed edits and retained starting points expand it further. Coverage is the fraction of the fixed scope--operation vocabulary reached by at least one valid edit; repeats add no coverage. It measures category coverage, not full space size or task quality, and does not separate ungenerated from invalid-only categories.

\textbf{Structural proposals and reuse.} Valid/attempted counts in Figure~\ref{fig:mechanism-plan}(c) are 9/14, 23/50, 22/45, 28/36, and 26/34. Structured Search achieves 77.8\% validity versus 48.9\% for Constrained Search, reducing unsuccessful structural proposals. Rates are undefined before any structural attempt. Figure~\ref{fig:mechanism-plan}(d) shows ten occupied categories and 17 cumulative archive-origin updates that pass incumbent admission. These retained versions supply used starting points; Section~\ref{sec:expected-cost} accounts for their overhead.

\end{multicols}
\par\addvspace{8pt}\noindent
\begin{minipage}{\textwidth}

\centering
\includegraphics[width=.84\textwidth]{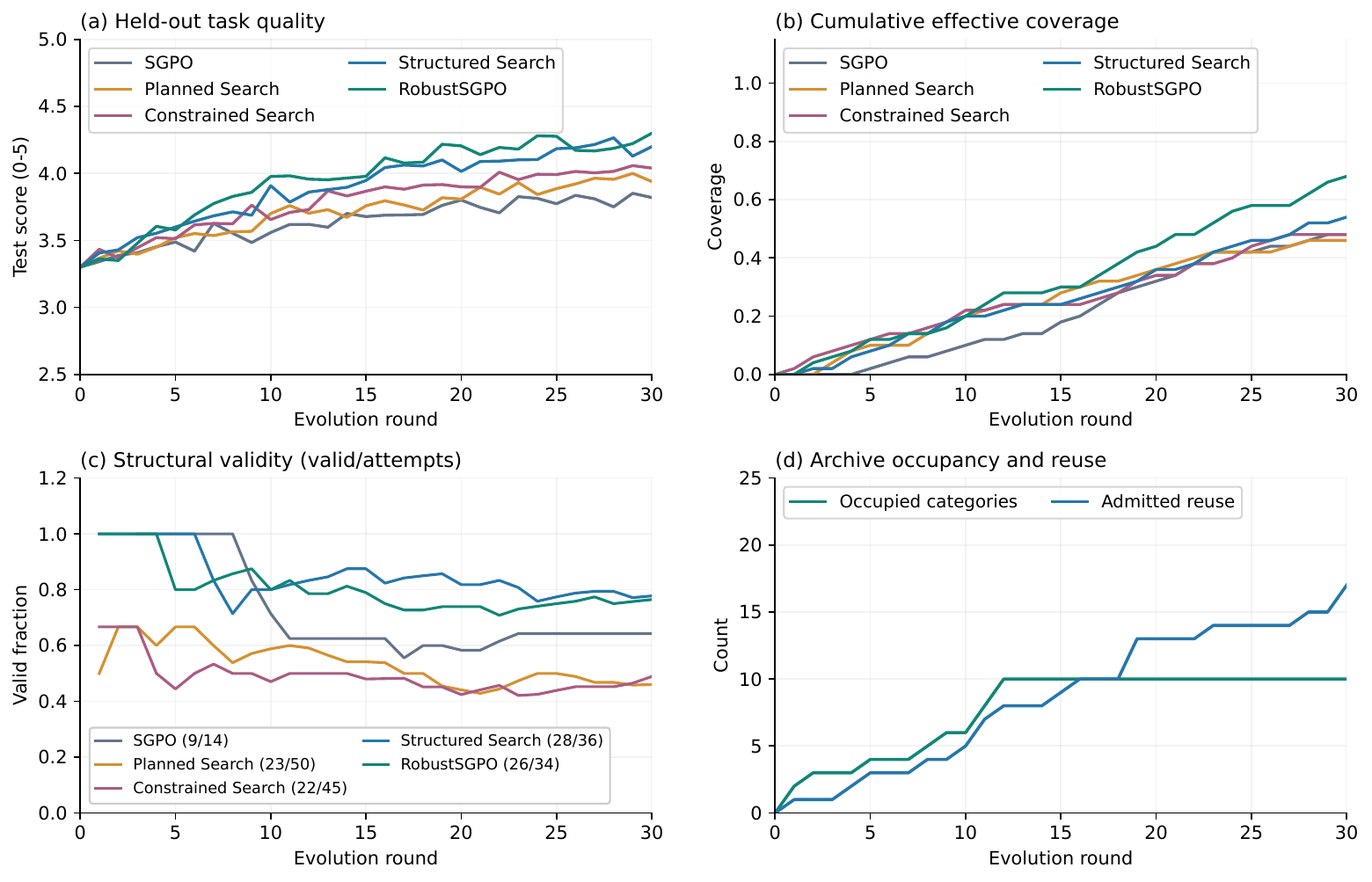}
\captionof{figure}{E2 search behavior: (a) held-out quality, (b) cumulative effective-category coverage, (c) cumulative structural validity with valid/attempted counts, and (d) occupied archive categories and cumulative admitted archive-origin updates.}
\label{fig:mechanism-plan}

\end{minipage}
\par\addvspace{12pt}
\begin{multicols}{2}\raggedcolumns

\subsection{E3: Adaptation and Retention After a Task Shift}
We evaluate local-to-cross-agent failures and the reverse. Each run spends ten rounds on source tasks and ten on destination tasks, using the respective optimization and validation splits. Retained snapshots are rescored on the destination validation set before reuse. We compare SGPO, Structured Search, RobustSGPO, and RobustSGPO (Random Archive). Random retention uses the same capacity, parent-use frequency, rescore budget, and admission rule, but stores valid snapshots by reservoir sampling rather than category-wise quality.

\end{multicols}
\par\addvspace{8pt}\noindent
\begin{minipage}{\textwidth}

\centering
\includegraphics[width=.98\textwidth]{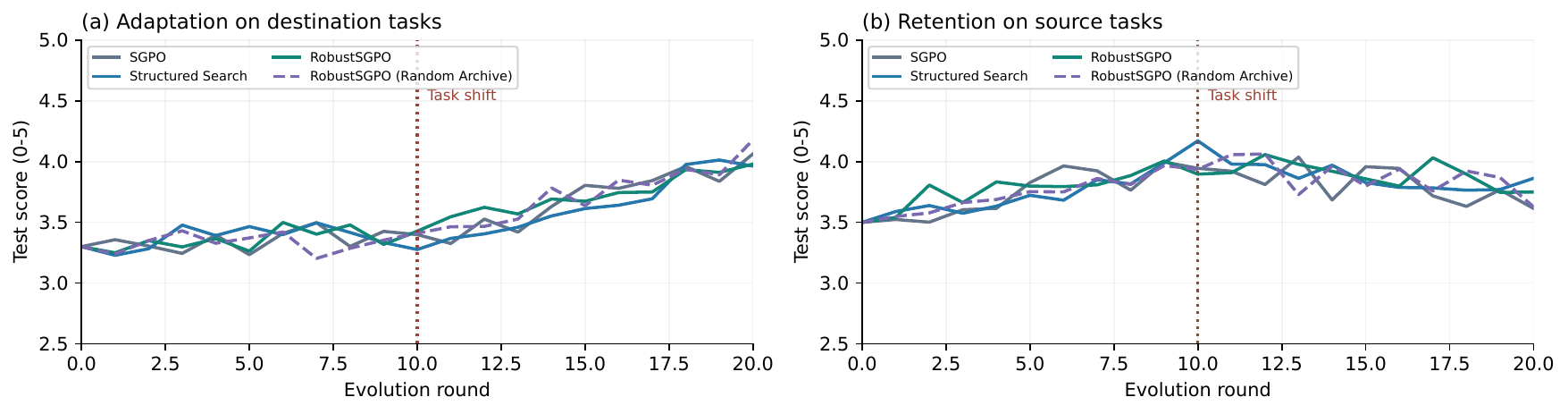}
\captionof{figure}{E3 adaptation and retention after the round-ten task shift. Dashed curves use random retention; category retention reduces source-task loss, while random retention reaches a higher destination endpoint.}
\label{fig:transfer-plan}

\end{minipage}
\par\addvspace{12pt}
\begin{multicols}{2}\raggedcolumns

Figure~\ref{fig:transfer-plan} tracks destination tasks on the left and source tasks on the right throughout the run. Mean destination scores over rounds 11--20 are 3.72, 3.67, 3.74, and 3.75; final scores are 4.06, 3.96, 3.98, and 4.18. RobustSGPO improves mean adaptation quality over Structured Search, while random retention obtains the highest destination endpoint.

Final source scores decline by 0.33, 0.31, 0.15, and 0.31 points relative to round ten. Category retention has the smallest source-task loss. Random retention improves destination quality further but loses more source capability. The archive policy therefore changes the adaptation--retention tradeoff: category elites benefit retention rather than maximizing the destination endpoint.

\end{multicols}
\par\addvspace{8pt}\noindent
\begin{minipage}{\textwidth}

\centering\small
\captionof{table}{Held-out task results and equal-token comparison. Local and cross-agent sets contain 15 tasks each. Round-30 scores correspond to Figure~\ref{fig:mechanism-plan}(a); the budget column uses the last affordable checkpoint within 20 million tokens. Bold marks column bests, including ties.}
\label{tab:expected-behavior}
\begin{tabular}{lrrrrr}
\toprule
Configuration & Round 30 & Local & Cross-agent & Success & 20M tokens\\
\midrule
SGPO & 3.82 & 11/15 & 7/15 & 60.0\% & 3.77\\
Planned Search & 3.94 & 11/15 & 8/15 & 63.3\% & 3.84\\
Constrained Search & 4.04 & 12/15 & 8/15 & 66.7\% & 3.99\\
Structured Search & 4.20 & 12/15 & \textbf{11/15} & 76.7\% & 4.10\\
RobustSGPO & \textbf{4.30} & \textbf{13/15} & \textbf{11/15} & \textbf{80.0\%} & \textbf{4.14}\\
\bottomrule
\end{tabular}

\end{minipage}
\par\addvspace{12pt}
\begin{multicols}{2}\raggedcolumns

\subsection{Behavioral Outcomes at Equal Token Cost}\label{sec:expected-cost}
Table~\ref{tab:expected-behavior} evaluates actual task completion, required artifacts, protected operations, and cross-agent handoffs. RobustSGPO completes 24/30 tasks versus SGPO's 18/30: local completion rises from 11/15 to 13/15, and cross-agent completion from 7/15 to 11/15.

\par\medskip\noindent\begin{minipage}{\linewidth}

\centering
\includegraphics[width=\linewidth]{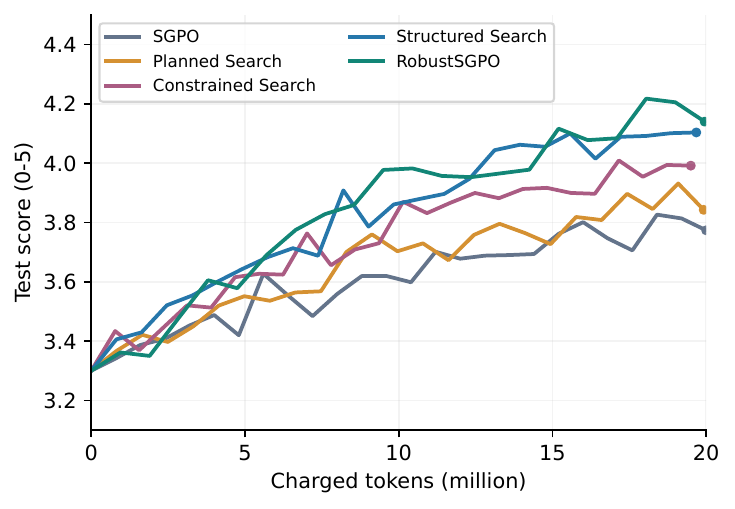}
\captionof{figure}{Test quality versus cumulative charged tokens, including all search-side model calls. Dots mark the last affordable checkpoints within 20 million tokens.}
\label{fig:equal-cost}
\end{minipage}\par\medskip

Per-round costs in Figure~\ref{fig:equal-cost} are 0.80, 0.83, 0.78, 0.82, and 0.95 million in table order. Within 20 million tokens, the final affordable rounds are 25, 24, 25, 24, and 21, yielding scores of 3.77, 3.84, 3.99, 4.10, and 4.14. RobustSGPO exceeds SGPO by 0.37 points at equal cost. Its margin over Structured Search shrinks from 0.10 at equal rounds to 0.04 at equal token cost, reflecting archive overhead. Checkpoints are selected by budget, not by test score.

\section{Conclusion}
RobustSGPO decomposes free-form SGPO updates into requested edits, patch construction and checking, and retained search starting points. Periodic permissions improve final quality, predefined add/remove operations improve structural validity, and the full method outperforms free search at equal token cost. Category retention reduces source-task degradation after transfer while adding search overhead. These results support explicit control of edit choices and search starting points in multi-agent harness evolution.

\bibliographystyle{plainnat}
\bibliography{references}

\begin{thebibliography}{15}
\providecommand{\natexlab}[1]{#1}
\providecommand{\url}[1]{\texttt{#1}}
\expandafter\ifx\csname urlstyle\endcsname\relax
  \providecommand{\doi}[1]{doi: #1}\else
  \providecommand{\doi}{doi: \begingroup \urlstyle{rm}\Url}\fi

\bibitem[Agrawal et~al.(2025)Agrawal, Tan, Soylu, Ziems, Khare, Opsahl-Ong,
  Singhvi, Shandilya, Ryan, Jiang, Potts, Sen, Dimakis, Stoica, Klein, Zaharia,
  and Khattab]{gepa2025}
Lakshya~A Agrawal, Shangyin Tan, Dilara Soylu, Noah Ziems, Rishi Khare, Krista
  Opsahl-Ong, Arnav Singhvi, Herumb Shandilya, Michael~J Ryan, Meng Jiang,
  Christopher Potts, Koushik Sen, Alexandros~G. Dimakis, Ion Stoica, Dan Klein,
  Matei Zaharia, and Omar Khattab.
\newblock {GEPA: Reflective Prompt Evolution Can Outperform Reinforcement
  Learning}, 2025.

\bibitem[Fan and Huang(2026)]{helix2026}
Tianyu Fan and Chao Huang.
\newblock {HELIX: Model-Harness Co-evolution for Recursive Self-Improvement},
  2026.

\bibitem[Fernando et~al.(2023)Fernando, Banarse, Michalewski, Osindero, and
  Rockt{\"a}schel]{promptbreeder2023}
Chrisantha Fernando, Dylan Banarse, Henryk Michalewski, Simon Osindero, and Tim
  Rockt{\"a}schel.
\newblock Promptbreeder: Self-referential self-improvement via prompt
  evolution, 2023.

\bibitem[Lao et~al.(2026)Lao, Pan, Ma, Li, Lin, Shi, et~al.]{agentx2026}
Changxin Lao, Fei Pan, Guozhuang Ma, Han Li, Huihuang Lin, Jijun Shi, et~al.
\newblock {AgentX: Towards Agent-Driven Self-Iteration of Industrial
  Recommender Systems}.
\newblock \emph{arXiv preprint arXiv:2606.26859}, 2026.
\newblock \doi{10.48550/arXiv.2606.26859}.
\newblock URL \url{https://arxiv.org/abs/2606.26859}.

\bibitem[Lee et~al.(2026)Lee, Nair, Zhang, Lee, Khattab, and
  Finn]{metaharness2026}
Yoonho Lee, Roshen Nair, Qizheng Zhang, Kangwook Lee, Omar Khattab, and Chelsea
  Finn.
\newblock {Meta-Harness: End-to-End Optimization of Model Harnesses}, 2026.

\bibitem[Lin et~al.(2026)Lin, Liu, Pan, Lin, Dou, Xi, Huang, Yan, Han, Gui, and
  Jiang]{ahe2026}
Jiahang Lin, Shichun Liu, Chengjun Pan, Lizhi Lin, Shihan Dou, Zhiheng Xi,
  Xuanjing Huang, Hang Yan, Zhenhua Han, Tao Gui, and Yu-Gang Jiang.
\newblock {Agentic Harness Engineering: Observability-Driven Automatic
  Evolution of Coding-Agent Harnesses}, 2026.

\bibitem[Liu et~al.(2026{\natexlab{a}})Liu, Shou, Liu, Wen, Chen, Fang, and
  Feng]{agentflow2026}
Hanzhi Liu, Chaofan Shou, Xiaonan Liu, Hongbo Wen, Yanju Chen, Ryan~Jingyang
  Fang, and Yu~Feng.
\newblock {Synthesizing Multi-Agent Harnesses for Vulnerability Discovery},
  2026{\natexlab{a}}.

\bibitem[Liu et~al.(2026{\natexlab{b}})Liu, Shi, Sang, He, Lin, Wei, Wang,
  Dumoulin, Jin, and Lu]{adaptiveharness2026}
Zewen Liu, Zhan Shi, Yisi Sang, Bing He, Minhua Lin, Tianxin Wei, Dakuo Wang,
  Benoit Dumoulin, Wei Jin, and Hanqing Lu.
\newblock {Adaptive Auto-Harness: Sustained Self-Improvement for Agentic System
  Deployment on Open-Ended Task Streams}, 2026{\natexlab{b}}.

\bibitem[Luo et~al.(2026)Luo, Huang, Luo, Liu, Li, Hu, Feng, and Liu]{hase2026}
Haochen Luo, Yi~Huang, Sichun Luo, Fengyuan Liu, Lei Li, Zefa Hu, Junlan Feng,
  and Qi~Liu.
\newblock {Harness-Aware Self-Evolving: Co-Evolving Model Weights, Harness, and
  Task Solutions}, 2026.

\bibitem[Mouret and Clune(2015)]{mapelites2015}
Jean-Baptiste Mouret and Jeff Clune.
\newblock Illuminating search spaces by mapping elites, 2015.

\bibitem[Sengupta and Wang(2026)]{harbor2026}
Biswa Sengupta and Jinhua Wang.
\newblock {HARBOR: Automated Harness Optimization}, 2026.

\bibitem[Wei et~al.(2026)Wei, Shi, Lin, He, Liu, Sang, Bei, Ning, Zou, Li, Lin,
  Zhao, Wang, Dumoulin, Wang, He, and Lu]{evoharness2026}
Tianxin Wei, Zhan Shi, Minhua Lin, Bing He, Zewen Liu, Yisi Sang, Yuanchen Bei,
  Xuying Ning, Jiaru Zou, Ting-Wei Li, Xiao Lin, Yanjun Zhao, Chi Wang, Benoit
  Dumoulin, Dakuo Wang, Jingrui He, and Hanqing Lu.
\newblock {Evo-Harness: Context-to-Harness Skill Compilation for Self-Evolving
  Agents}, 2026.

\bibitem[Yang et~al.(2023)Yang, Wang, Lu, Liu, Le, Zhou, and Chen]{opro2023}
Chengrun Yang, Xuezhi Wang, Yifeng Lu, Hanxiao Liu, Quoc~V. Le, Denny Zhou, and
  Xinyun Chen.
\newblock Large language models as optimizers, 2023.

\bibitem[Yuksekgonul et~al.(2024)Yuksekgonul, Bianchi, Boen, Liu, Huang,
  Guestrin, and Zou]{textgrad2024}
Mert Yuksekgonul, Federico Bianchi, Joseph Boen, Sheng Liu, Zhi Huang, Carlos
  Guestrin, and James Zou.
\newblock Textgrad: Automatic ``differentiation'' via text, 2024.

\bibitem[Zhang et~al.(2026)Zhang, Zhang, Li, Zhang, Chen, Zhang, Bai, and
  Hu]{selfharness2026}
Hangfan Zhang, Shao Zhang, Kangcong Li, Chen Zhang, Yang Chen, Yiqun Zhang, Lei
  Bai, and Shuyue Hu.
\newblock {Self-Harness: Harnesses That Improve Themselves}, 2026.

\end{thebibliography}
\end{multicols}
\end{document}